\documentclass[runningheads]{llncs}

\usepackage[T1]{fontenc}
\usepackage{amsmath,amssymb}
\usepackage{booktabs}
\usepackage{algorithm}
\usepackage{algpseudocode}
\usepackage{graphicx}
\usepackage{xcolor}
\usepackage{url}
\usepackage{marvosym}

\newcommand{\safeplotplaceholder}{%
    \fbox{\parbox[c][0.55\linewidth][c]{0.92\linewidth}{\centering Missing figure}}%
}
\newcommand{\safeincludegraphics}[2][]{%
    \IfFileExists{#2}{\includegraphics[#1]{#2}}{\safeplotplaceholder}%
}
\newcommand{\safeplot}[2][]{%
    \IfFileExists{#2}{\includegraphics[#1]{#2}}{\safeplotplaceholder}%
}

\title{Private Synthetic Data Generation for Continuous Data via Adaptive Binning and K-Means}
\titlerunning{Private Synthetic Data for Continuous Data}

\author{Koki Kugimiya\inst{2} \and Papangkorn Inkeaw\inst{2} \and Vorapong Suppakitpaisarn\inst{1}\textsuperscript{\Letter}}
\authorrunning{K. Kugimiya et al.}
\institute{Graduate School of Information Science and Technology, The University of Tokyo, Tokyo, Japan\\
\email{vorapong@is.s.u-tokyo.ac.jp}
\and
Data Science Research Center, Faculty of Science, Chiang Mai University, Chiang Mai, Thailand\\
\email{kokikungimiya@gmail.com, papangkorn.i@cmu.ac.th}}

\begin{document}
\maketitle

\begin{abstract}
PrivMRF is a standard method for synthesizing data under differential privacy. However, PrivMRF assumes that input data are categorical, and therefore continuous attributes must be discretized before the method can be applied. A common approach for applying PrivMRF to continuous data is equal-width binning, in which all bins have the same interval size. This approach can incur large information loss when the data distribution is skewed, because many bins contain only a very small number of records. To address this limitation, we propose two discretization methods. The first method is adaptive binning. Each continuous attribute is first divided into very small bins; the noisy frequency of each small bin is then published, and adjacent small bins are merged until each resulting bin has adequate frequency. The second method is based on K-Means. We employ differentially private K-Means to obtain $K$ cluster centers, and define bins according to the range of values closest to each center. The experimental results showed that distribution-aware discretization substantially improved utility over equal-width binning, especially under strict privacy constraints: at privacy loss $0.2$, adaptive CDF binning improved accuracy from 0.6345 to 0.7158 on Avila dataset and from 0.9144 to 0.9594 on MAGIC dataset. On Dry Bean dataset at privacy loss $0.1$, differentially private K-Means increased macro-F1 from 0.0651 to 0.8718.
\end{abstract}

\section{Introduction}

Differential privacy (DP) provides a rigorous framework for limiting the disclosure of any individual record in a released output~\cite{dwork2006,dwork2014}. In machine learning and data analysis, this guarantee has been incorporated into empirical risk minimization, differentially private stochastic gradient descent (DP-SGD), and teacher--student frameworks such as PATE~\cite{chaudhuri2011,abadi2016,papernot2018}. These methods protect privacy by injecting calibrated randomness, but the injected noise creates a fundamental trade-off between privacy and utility.

Differentially private synthetic data is one practical way to manage this trade-off. A synthetic dataset can be released once under a fixed privacy guarantee, and subsequent analyses that use only the released synthetic records do not consume additional privacy budget by the post-processing property of DP~\cite{dwork2014}. This makes DP synthetic data attractive when researchers need shareable, data-like records for repeated downstream analysis while limiting disclosure risk for the original individuals~\cite{mckenna2019,10.14778/3476249.3476272}.

For high-dimensional tabular data, marginal-based graphical models are a particularly important family of DP synthesizers. PrivMRF is a representative example: instead of estimating the full joint distribution directly, it selects informative low-dimensional marginals, perturbs them under DP, and combines them through a Markov random field~\cite{10.14778/3476249.3476272}. This design is attractive because it gives a stable and interpretable alternative to neural generators whose performance can depend heavily on architecture and optimization choices. PrivMRF has also influenced later marginal-based systems, including AIM, PrivLava, and VertiMRF~\cite{mckenna2022aim,cai2023privlava,zhao2024vertimrf}.

Continuous DP synthetic-data generation is not an unexplored problem. Prior work includes copula-based synthesis such as DPCopula, GAN-based methods such as PATE-GAN and DP-CTGAN, and kernel-embedding approaches such as DP-MERF~\cite{li2014dpcopula,jordon2019pategan,fang2022dpctgan,harder2021dpmerf}. These methods show that continuous or mixed tabular data can be synthesized under DP. However, they do not solve the distinct problem studied here: how to adapt continuous data to PrivMRF-style marginal/MRF synthesis. PrivMRF's measurement selection, noisy marginal estimation, and graphical-model inference are defined over finite categorical domains. Therefore, when PrivMRF is chosen for its marginal-based structure and robustness, continuous attributes cannot simply be passed to the synthesizer, nor is replacing PrivMRF with a generic continuous synthesizer the same objective.

This paper addresses that missing interface between continuous attributes and marginal-based MRF synthesis. In this setting, discretization is not merely a preprocessing detail: it determines the categorical domains on which noisy marginals are measured, controls sparsity in those marginals, and determines how synthetic discrete records can be reconstructed into continuous values. The common equal-width solution partitions each feature range into intervals of identical width, but it ignores the empirical density of the attribute. In skewed distributions, many bins may contain very few records, so DP noise dominates their counts, while dense regions remain too coarse. Useful continuous structure can therefore be lost before PrivMRF begins private synthesis.

\subsection{Our Contribution}

\begin{figure}[t]
    \centering
    \includegraphics[width=0.95\textwidth]{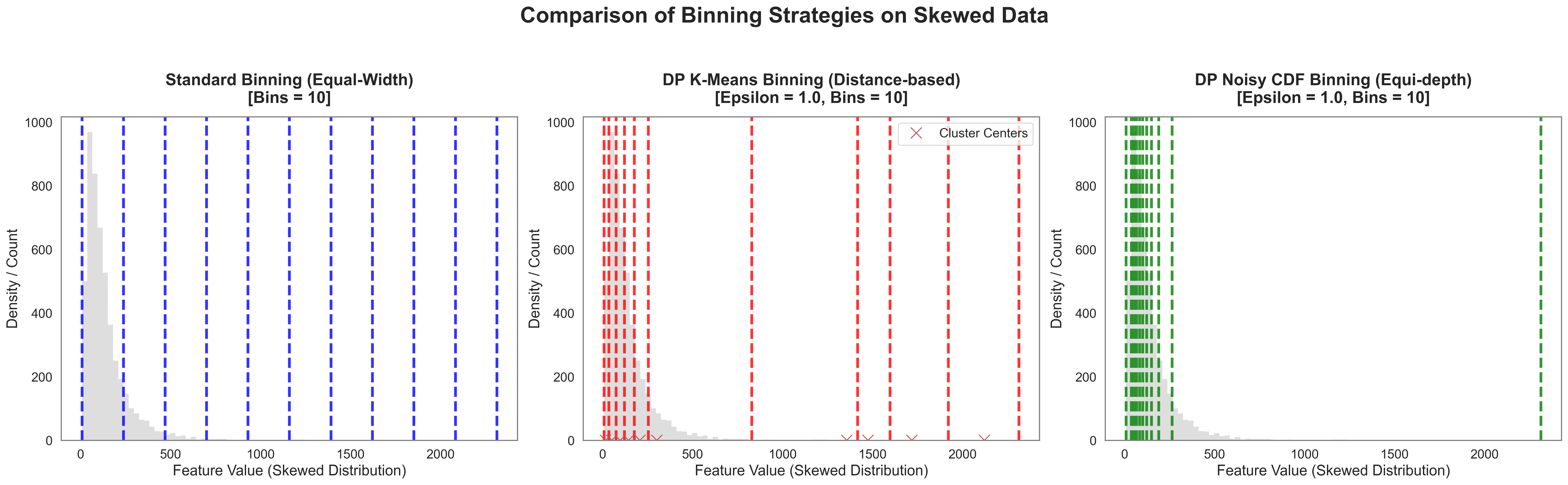}
    \caption{Our distribution-aware discretizations for applying PrivMRF to continuous data.}
    \label{fig:motivation_overview}
\end{figure}

As shown in Fig.~\ref{fig:motivation_overview}, we propose a private, distribution-aware domain interface that enables PrivMRF to synthesize continuous tabular data while keeping its marginal-based MRF core unchanged. The individual ingredients we use, noisy histograms/CDFs and DP K-Means, are standard DP tools; the contribution is to organize them into an end-to-end, privacy-accounted discretization and reconstruction pipeline for PrivMRF-style synthesis, where the learned bins must be compact enough for noisy marginal estimation and informative enough for downstream continuous-data utility.

The first proposed interface is noisy-CDF adaptive binning. It estimates a private marginal cumulative distribution function from noisy micro-bin counts, stabilizes the estimate through smoothing and isotonic regression, and selects adaptive boundaries from quantile-like levels. This gives narrower bins in dense regions and wider bins in sparse regions while keeping the final PrivMRF domain compact. The second interface is DP K-Means binning. For each continuous attribute, it runs one-dimensional differentially private K-Means to obtain private representative centers; these centers define both the discrete identifiers used by PrivMRF and the representative values used to reconstruct synthetic records. To improve low-budget stability, the method uses a budget-adaptive iteration limit.

Both interfaces are applied independently to each continuous attribute before PrivMRF synthesis. After PrivMRF generates synthetic discrete records, the generated bin identifiers are mapped back to continuous representative values for downstream analysis. We also state the privacy guarantee of the full pipeline explicitly: bin learning and PrivMRF compose to the total privacy budget, and reconstruction is post-processing of differentially private outputs.

Experiments on Dry Bean, MAGIC Gamma Telescope, and Avila show that treating discretization as a private domain-interface mechanism can substantially improve train-synthetic, test-real utility over equal-width binning, especially under strict privacy budgets. At $\epsilon=0.2$, adaptive CDF binning improves accuracy from 0.6345 to 0.7158 on Avila and from 0.9144 to 0.9594 on MAGIC. On Dry Bean at $\epsilon=0.1$, DP K-Means increases macro-F1 from 0.0651 to 0.8718, indicating that adaptive discretization can preserve class-level structure even when the privacy budget is very small.

\section{Background}

\paragraph{Differential privacy} Differential privacy (DP) provides a rigorous mathematical framework for quantifying and bounding the privacy risks associated with statistical data releases~\cite{dwork2006,dwork2014}. A randomized algorithm or mechanism $\mathcal{M}$ satisfies $\epsilon$-differential privacy if for all neighboring datasets $D$ and $D'$ differing by at most one record, and for all subsets of possible outputs $S \subseteq \text{Range}(\mathcal{M})$, 
$\Pr[\mathcal{M}(D) \in S] \le e^\epsilon \Pr[\mathcal{M}(D') \in S]$.
The parameter $\epsilon$ (epsilon), often referred to as the privacy budget, controls the strictness of the privacy guarantee. The composition theorem of DP states that when several DP mechanisms are applied to the same sensitive dataset, their privacy losses accumulate; in the basic case, mechanisms with privacy losses $\epsilon_1,\ldots,\epsilon_k$ provide an overall guarantee of at most $\sum_{i=1}^k \epsilon_i$-DP. %This property is important in our setting because the total privacy budget must be shared between the discretization step and the subsequent PrivMRF synthesis step.

\paragraph{Post-processing property}
Differential privacy is closed under post-processing: if a mechanism $\mathcal{M}$ satisfies $\epsilon$-DP, then any randomized or deterministic function applied only to its output also satisfies $\epsilon$-DP~\cite{dwork2014}. Therefore, once a synthetic dataset has been generated by a DP mechanism, subsequent analyses performed only on the released synthetic data do not consume additional privacy budget. 

\paragraph{Laplace mechanism}
The Laplace mechanism is a standard way to make numeric queries differentially private. For a vector-valued query $f(D)$ with $\ell_1$ sensitivity
    $\Delta_1(f)=\max_{D\sim D'}\|f(D)-f(D')\|_1$,
the mechanism releases
\begin{equation}
    \mathcal{M}(D)=f(D)+(\eta_1,\ldots,\eta_m), \qquad
    \eta_r \sim \mathrm{Laplace}\left(0,\frac{\Delta_1(f)}{\epsilon}\right),
\end{equation}
where independent Laplace noise is added to each coordinate~\cite{dwork2014}. %Thus, smaller $\epsilon$ or larger sensitivity produces larger perturbations. In our setting, this mechanism is used for the noisy count and sum statistics required by discretization: histogram counts have sensitivity one under add/remove neighboring datasets, while bounded coordinate sums have sensitivity determined by the public feature range.

\paragraph{The PrivMRF Architecture}
PrivMRF is a differentially private synthetic-data generator for high-dimensional tabular data~\cite{10.14778/3476249.3476272}. Its goal is to learn useful relationships among columns without releasing the original records. Instead of trying to estimate the full table distribution directly, which becomes infeasible when many attributes are present, PrivMRF selects a small set of informative low-dimensional summaries, such as one-column or two-column frequency tables. It then adds privacy noise to these summaries and combines them with a Markov random field (MRF), a graphical model that represents which attributes are statistically related.

After these noisy summaries are obtained, PrivMRF combines them into a single synthetic-data distribution using graphical-model inference. This step works best when the discretized attribute domains remain compact. Since PrivMRF expects categorical inputs, continuous values are commonly converted using equal-width binning, which divides each feature range into intervals of the same width. However, fixed-width bins can create many sparse categories, especially when several attributes are considered together; the resulting frequency tables are harder to estimate and the added DP noise has a larger effect. Our approach avoids this bottleneck by discretizing each continuous attribute independently in a distribution-aware way, keeping each attribute domain compact while still allowing PrivMRF's graphical model to learn dependencies across attributes after discretization.

\section{Proposed Method}

\subsection{Problem Setting and Privacy Budget}

Let $D = \{(\mathbf{x}_i, y_i)\}_{i=1}^n$ be a tabular dataset, where $\mathbf{x}_i \in \mathbb{R}^d$ contains $d$ continuous features and $y_i$ is a class label. The goal is to generate a synthetic dataset $D_{\mathrm{syn}}$ that preserves statistical and predictive utility while satisfying a total privacy budget $\epsilon$.

PrivMRF requires finite discrete domains. Therefore, each continuous feature $j$ is first mapped by a univariate encoder
    $q_j: [l_j,u_j] \rightarrow \{0,1,\ldots,B_j-1\}$,
where $[l_j,u_j]$ is a public feature range and $B_j$ is the number of target bins or representatives. The full encoded record is
    $Q(\mathbf{x}_i)=\big(q_1(x_{i1}),q_2(x_{i2}),\ldots,q_d(x_{id})\big)$,
and the encoded dataset is denoted by $D^{\mathrm{disc}}=\{(Q(\mathbf{x}_i),y_i)\}_{i=1}^n$.

We denote the private discretization mechanism by $M_{\mathrm{bin}}$ and the private PrivMRF synthesis mechanism by $M_{\mathrm{mrf}}$. Their privacy budgets are $\epsilon_{\mathrm{bin}}$ and $\epsilon_{\mathrm{mrf}}$, respectively: $\epsilon_{\mathrm{bin}}$ is used to learn private binning information, and $\epsilon_{\mathrm{mrf}}$ is used to synthesize records from the discretized data. By the basic     $\epsilon = \epsilon_{\mathrm{bin}} + \epsilon_{\mathrm{mrf}}$.
The binning budget is divided equally across the $d$ continuous features, so each feature receives
    $\epsilon_j = \frac{\epsilon_{\mathrm{bin}}}{d}$.
\subsection{Overall Pipeline}

The proposed framework consists of three stages. First, each continuous attribute is discretized independently using either DP K-Means binning or noisy-CDF adaptive binning. Second, the discretized records form the internal representation $D^{\mathrm{disc}}$, on which PrivMRF is trained to generate synthetic discrete records. Finally, the synthetic bin identifiers are mapped back to continuous values using the private representatives learned during discretization. The only public output is $D_{\mathrm{syn}}$; bin assignments, boundaries, centers, noisy counts, and noisy means remain internal to the composed mechanism.

\begin{theorem}[Privacy of the full pipeline]
Assume that each feature-level bin-learning mechanism $M_{\mathrm{bin},j}$ satisfies $\epsilon_j$-DP and that $\sum_{j=1}^{d}\epsilon_j=\epsilon_{\mathrm{bin}}$. Let $M_{\mathrm{bin}}(D)$ denote the joint bin-learning output, including all private boundaries, centers, noisy counts, and noisy means used for discretization and reconstruction. For any fixed output of $M_{\mathrm{bin}}$, the induced encoder maps neighboring raw datasets to neighboring discrete datasets. If PrivMRF, denoted by $M_{\mathrm{mrf}}$, satisfies $\epsilon_{\mathrm{mrf}}$-DP on neighboring discrete datasets, then the complete mechanism
\begin{equation}
    \mathcal{A}(D)
    = R\!\left(M_{\mathrm{bin}}(D),
    M_{\mathrm{mrf}}\left(Q_{M_{\mathrm{bin}}(D)}(D)\right)\right)
\end{equation}
that learns bins, synthesizes discrete records with PrivMRF, and reconstructs continuous synthetic records satisfies
    $(\epsilon_{\mathrm{bin}}+\epsilon_{\mathrm{mrf}})\text{-DP}
    = \epsilon\text{-DP}$.
\end{theorem}

\subsection{DP K-Means Binning}

Algorithm~\ref{alg:kmeans} summarizes the DP K-Means binning stage and the reconstruction rule applied after PrivMRF generates synthetic bin identifiers.

\begin{algorithm}[t]
\caption{Univariate DP K-Means binning and reconstruction with dynamic iteration limit}
\label{alg:kmeans}
\begin{algorithmic}[1]
\Require Continuous feature $X_j$, bins $B_j$, total binning budget $\epsilon_{\mathrm{bin}}$, features $d$, bounds $[l_j,u_j]$
\Ensure Internal discrete codes $Z_j$, private centers $\{c_{j,k}\}_{k=0}^{B_j-1}$, and reconstructed synthetic feature $\widetilde{X}_j$
\State Set per-feature budget $\epsilon_j \gets \epsilon_{\mathrm{bin}}/d$
\State Set $\epsilon_{\mathrm{kmeans}} \leftarrow \epsilon_j$
\If{$\epsilon_{\mathrm{kmeans}} \le 0.8$}
    \State Set maximum iterations $T \gets 1$
\Else
    \State Set $T \gets T_{\mathrm{default}}$
\EndIf
\State Fit one-dimensional DP K-Means with $B_j$ clusters, $T$ iterations, and budget $\epsilon_{\mathrm{kmeans}}$
\State Sort private centers $c_{j,0},\ldots,c_{j,B_j-1}$
\State Encode each value as $z_{ij}\gets \arg\min_{0\le k<B_j}|x_{ij}-c_{j,k}|$
\State Form the internal discrete column $Z_j=(z_{1j},\ldots,z_{nj})$
\State After PrivMRF returns synthetic bin identifiers $\widetilde{Z}_j$, reconstruct each value as $\widetilde{x}_{ij}\gets c_{j,\widetilde{z}_{ij}}$
\State \Return $\widetilde{X}_j$ as part of $D_{\mathrm{syn}}$; keep $Z_j$ and $\{c_{j,k}\}$ internal
\end{algorithmic}
\end{algorithm}

For a feature $X_j = (x_{1j},\ldots,x_{nj})$, DP K-Means fits $B_j$ private cluster centers within the public bounds $[l_j,u_j]$. Each value is encoded by the index of the nearest private center, producing a discrete column suitable for PrivMRF. The full per-feature budget is passed to the DP Lloyd routine~\cite{su2016} with
    $\epsilon_{\mathrm{kmeans}} = \epsilon_j = \frac{\epsilon_{\mathrm{bin}}}{d}$.
The DP Lloyd routine injects calibrated noise into the count and coordinate-sum statistics used in centroid updates.

\paragraph{Dynamic iteration limit.}
Repeated noisy centroid updates can reduce utility when $\epsilon_{\mathrm{kmeans}}$ is small. To avoid unstable center movement in the severe-privacy regime, we restrict the number of Lloyd iterations according to
\begin{equation}
    T =
    \begin{cases}
    1, & \epsilon_{\mathrm{kmeans}} \le 0.8,\\
    T_{\mathrm{default}}, & \epsilon_{\mathrm{kmeans}} > 0.8.
    \end{cases}
\end{equation}
The threshold is expressed on the per-feature budget scale. At low budgets, using one update preserves a coarse but stable partition of the marginal distribution; at higher budgets, the default iteration count is used. Experiments across the tested thresholds showed that $0.8$ gave the best overall utility.

\paragraph{Centroids as bins.}
After the final Lloyd iteration, the private centroids are sorted as
    $c_{j,0} \le c_{j,1} \le \cdots \le c_{j,B_j-1}$.
They then induce one-dimensional Voronoi bins: each boundary is placed halfway between two adjacent centroids,
\begin{equation}
    b_{j,k}=\frac{c_{j,k-1}+c_{j,k}}{2}, \qquad k=1,\ldots,B_j-1,
\end{equation}
with the public endpoints $l_j$ and $u_j$ closing the first and last intervals. Each original value is assigned to the bin whose centroid is nearest, so the Lloyd centroids define both the discrete bin identifiers used by PrivMRF and the representative values used later for reconstruction.

\paragraph{Reconstruction.}
For continuous data reconstruction, each synthetic bin identifier generated by PrivMRF is replaced by its corresponding private cluster center. For a synthetic bin code $z_j \in \{0,\ldots,B_j-1\}$,
    $\tilde{x}_j = c_{j,z_j}$,
where $c_{j,z_j}$ is the private centroid of cluster $z_j$. Algorithm~\ref{alg:kmeans} outputs this reconstructed synthetic feature as part of $D_{\mathrm{syn}}$, while the discrete codes and centroid table remain internal.

\subsection{Noisy-CDF Adaptive Binning}

\begin{algorithm}[t]
\caption{DP noisy-CDF adaptive binning and reconstruction with isotonic regression}
\label{alg:cdf}
\begin{algorithmic}[1]
\Require Continuous feature $X_j$, micro-bins $M=2000$, target bins $B_j$, feature budget $\epsilon_j$, bounds $[l_j,u_j]$
\Ensure Internal codes $Z_j$, intervals $\{I_{j,k}\}$, private bin centers $\{\hat{\mu}_{j,k}\}$, and reconstructed feature $\widetilde{X}_j$
\State Set $\epsilon_{\mathrm{hist}}\gets0.5\epsilon_j$, $\epsilon_{\mathrm{sum}}\gets0.25\epsilon_j$, and $\epsilon_{\mathrm{count}}\gets0.25\epsilon_j$
\State Compute micro-bin histogram $h_1,\ldots,h_M$
\State Add Laplace noise with scale $\Delta_{\mathrm{hist}}/\epsilon_{\mathrm{hist}}$ to obtain $\hat{h}_1,\ldots,\hat{h}_M$
\State Smooth noisy counts using a convolution kernel
\State Accumulate smoothed counts into a noisy CDF $\hat{F}$
\State Apply isotonic regression to obtain monotone CDF $g$
\State Clip and normalize $g$; if the noisy total is nonpositive, use uniform fallback boundaries
\State Select $B_j$ equal-frequency intervals $I_{j,0},\ldots,I_{j,B_j-1}$ from the generalized inverse of $g$ at levels $k/B_j$, retaining empty codes when boundaries duplicate
\State Encode each value as $z_{ij}$ according to its adaptive interval
\State Compute private bin centers $\hat{\mu}_{j,k}$ using shifted-sum budget $\epsilon_{\mathrm{sum}}$ and count budget $\epsilon_{\mathrm{count}}$
\State Form the internal discrete column $Z_j=(z_{1j},\ldots,z_{nj})$
\State After PrivMRF returns synthetic bin identifiers $\widetilde{Z}_j=(\widetilde{z}_{1j},\ldots,\widetilde{z}_{n_{\mathrm{syn}}j})$, reconstruct each value as $\widetilde{x}_{ij}\gets \hat{\mu}_{j,\widetilde{z}_{ij}}$
\State \Return $\widetilde{X}_j$ as part of $D_{\mathrm{syn}}$; keep $Z_j$, intervals, and private centers internal
\end{algorithmic}
\end{algorithm}

As summarized in Algorithm~\ref{alg:cdf}, the second discretizer constructs adaptive equal-frequency bins from a private estimate of the marginal cumulative distribution function (CDF). Unlike DP K-Means, which learns representative centers directly, this method estimates the one-dimensional distribution shape and selects bin boundaries from private quantile information.

\paragraph{Private histogram and budget split.}
For each feature $X_j$, the domain $[l_j,u_j]$ is divided into $M$ micro-bins; in our implementation, $M=2000$. Let $h_m$ be the count in micro-bin $m$. The feature-level budget is divided into three parts: $\epsilon_{\mathrm{hist}}$ for perturbing the micro-bin histogram used to estimate the CDF, $\epsilon_{\mathrm{sum}}$ for perturbing shifted interval sums used in reconstruction, and $\epsilon_{\mathrm{count}}$ for perturbing interval counts used in reconstruction. Thus,
\begin{equation}
    \epsilon_j = \epsilon_{\mathrm{hist}} + \epsilon_{\mathrm{sum}} + \epsilon_{\mathrm{count}},
\end{equation}
with the default allocation
\begin{equation}
    \epsilon_{\mathrm{hist}}=0.5\epsilon_j,
    \qquad
    \epsilon_{\mathrm{sum}}=0.25\epsilon_j,
    \qquad
    \epsilon_{\mathrm{count}}=0.25\epsilon_j.
\end{equation}
We use add/remove neighboring datasets, so the micro-bin histogram has $\ell_1$ sensitivity $\Delta_{\mathrm{hist}}=1$. The noisy histogram is
\begin{equation}
    \hat{h}_m = h_m + \mathrm{Laplace}\left(\frac{\Delta_{\mathrm{hist}}}{\epsilon_{\mathrm{hist}}}\right).
\end{equation}
The noisy counts are accumulated to form a noisy CDF. Under small $\epsilon_{\mathrm{hist}}$, however, independent noise can create local inversions and irregular jumps, which makes quantile boundaries unstable.

\paragraph{Smoothing and monotone repair.}
To reduce local perturbations, the noisy micro-bin counts are first smoothed using a convolution kernel~\cite{parzen1962}. We then apply isotonic regression~\cite{barlow1972} to the cumulative curve:
\begin{equation}
    \min_{g_1 \leq g_2 \leq \cdots \leq g_M} \sum_{m=1}^{M} (g_m - \hat{F}_m)^2,
\end{equation}
where $\hat{F}_m$ is the smoothed noisy CDF value at micro-bin $m$. The fitted curve is clipped to $[0,1]$ and normalized when the noisy total is positive; if the noisy total is nonpositive, the mechanism falls back to uniform boundaries over $[l_j,u_j]$. This monotone repair prevents bin boundaries from crossing or oscillating and makes adaptive bins more stable under strong privacy noise.

\paragraph{Boundary selection and reconstruction.}
After obtaining the monotone CDF, the algorithm selects $B_j$ target bins using equal-frequency cut points at CDF levels $0,1/B_j,2/B_j,\ldots,1$. The resulting nonempty intervals are narrower in dense regions and wider in sparse regions. The final PrivMRF domain remains $\{0,\ldots,B_j-1\}$; if adjacent target levels map to the same micro-bin, the corresponding empty code uses a deterministic midpoint fallback during reconstruction.

To reconstruct continuous values, each synthetic code generated by PrivMRF is published as the private representative center of its adaptive bin. In our implementation, this center is the noisy bin mean:
\begin{equation}
    \tilde{x}_j = \hat{\mu}_{j,z_j}.
\end{equation}
The representative center is computed from a noisy shifted interval sum and a noisy interval count:
\begin{equation}
    \hat{s}'_{j,z_j}=s'_{j,z_j}+\mathrm{Laplace}\left(\frac{\Delta_{\mathrm{sum}}}{\epsilon_{\mathrm{sum}}}\right),\qquad
    \hat{n}_{j,z_j}=n_{j,z_j}+\mathrm{Laplace}\left(\frac{\Delta_{\mathrm{count}}}{\epsilon_{\mathrm{count}}}\right),
\end{equation}
where $s'_{j,z_j}=\sum_{i:q_j(x_{ij})=z_j}(x_{ij}-l_j)$, $\Delta_{\mathrm{sum}}=u_j-l_j$, and $\Delta_{\mathrm{count}}=1$. If $\hat{n}_{j,z_j}>0$, the representative center is $l_j+\hat{s}'_{j,z_j}/\hat{n}_{j,z_j}$ clamped to the selected interval; otherwise, it falls back to the interval midpoint. Algorithm~\ref{alg:cdf} outputs this reconstructed synthetic feature as part of $D_{\mathrm{syn}}$, while the discrete codes, intervals, and private centers remain internal.

\subsection{Integration with PrivMRF}

After the continuous features are converted into bins, PrivMRF is trained on the discretized dataset $D^{\mathrm{disc}}$ and generates synthetic bin identifiers. Each synthetic bin identifier is then converted back into a continuous value by replacing it with the corresponding private representative: a DP K-Means centroid or a noisy-CDF bin center. In this way, the proposed method keeps the PrivMRF synthesizer unchanged and only changes how continuous features are represented before and after synthesis.

\section{Experimental Setup}

This section describes the experimental setup: datasets, baselines, privacy settings, and evaluation protocol.

\subsection{Datasets}

We evaluate the framework on three classification datasets with different feature dimensions and class structures. Table~\ref{tab:datasets} summarizes the main properties. Dry Bean tests a high-dimensional multiclass setting, MAGIC Gamma Telescope tests a binary task with smooth continuous features, and Avila tests a multiclass setting with overlapping or imbalanced classes. These datasets provide representative benchmarks for evaluating machine-learning methods under differential privacy. We focus on datasets with at most 16 features because the computational and statistical costs of many differentially private methods increase substantially with dimensionality. For every dataset, the continuous feature bounds $[l_j,u_j]$ are treated as public domain information fixed before inspecting the private training data. These bounds are reused across all folds and are not estimated from private records.

\begin{table}[t]
\centering
\caption{Datasets used in the experimental evaluation.}
\label{tab:datasets}
\begin{tabular}{lccp{4.5cm}}
\toprule
Dataset & Features & Classes & Main challenge \\
\midrule
Dry Bean & 16 & 7 & High-dimensional geometric morphology \\
MAGIC Gamma Telescope & 10 & 2 & Smooth continuous astrophysical signals \\
Avila & 10 & 12 & Multiclass imbalance and overlapping classes \\
\bottomrule
\end{tabular}
\end{table}

\subsection{Compared Methods}

We use a support vector machine (SVM)~\cite{cortes1995} as the downstream classifier throughout this work. For real-data evaluation, the SVM is trained and tested on real records. For synthetic-data evaluation, the SVM is applied to the reconstructed synthetic data: it is trained on synthetic records and tested on held-out real records. Since the SVM training uses only the released synthetic data, it is a post-processing step of the differentially private synthetic-data mechanism and therefore does not consume any additional privacy budget. 

We compare the proposed discretization front ends with two reference baselines. \textbf{Absolute TRTR} (train-real, test-real) is a non-private upper bound: the SVM is trained and tested on real data using the same 5-fold splits. It shows the best utility expected without privacy noise or synthetic-data generation. \textbf{Uniform-PrivMRF} is the private baseline: each continuous feature is discretized with equal-width bins before applying PrivMRF. It uses the same privacy-budget grid, bin-count grid, reconstruction protocol, and evaluation pipeline as the proposed methods. The two proposed private variants replace equal-width binning with either DP K-Means binning or noisy-CDF adaptive binning.

\subsection{Privacy and Discretization Settings}

All experiments use the total privacy budgets
\begin{equation}
    \epsilon \in \{0.1, 0.2, 0.4, 0.8, 1.6, 3.2\}.
\end{equation}
These values cover both strict-privacy and moderate-privacy regimes. The total budget is split between discretization and PrivMRF synthesis, with tested ratios
\begin{equation}
    \epsilon_{\mathrm{mrf}} : \epsilon_{\mathrm{bin}}
    \in \{9{:}1, 8{:}2, 7{:}3, 6{:}4, 5{:}5\}.
\end{equation}
In the result labels, the method name is followed by this budget split: for example, CDF~64 denotes noisy-CDF adaptive binning with $\epsilon_{\mathrm{mrf}}:\epsilon_{\mathrm{bin}}=6{:}4$, and KM~55 denotes DP K-Means binning with $\epsilon_{\mathrm{mrf}}:\epsilon_{\mathrm{bin}}=5{:}5$.
We keep at least half of the total budget for PrivMRF, i.e., $\epsilon_{\mathrm{mrf}} \ge 0.5\epsilon$, because smaller PrivMRF budgets produced severe utility loss in graphical-model estimation and inference. Across the tested splits, the equal $5{:}5$ allocation gave the best overall utility and is therefore the default.

For both proposed discretizers, the target number of bins per feature is varied from $B_j=2$ to $B_j=21$. The upper limit is chosen because predictive utility empirically plateaued beyond this range. %In the noisy-CDF method, each feature is first divided into $M=2000$ micro-bins before adaptive merging. Within the binning budget, each feature receives $\epsilon_j=\epsilon_{\mathrm{bin}}/d$. DP K-Means uses this feature budget as $\epsilon_{\mathrm{kmeans}}$, with the one-iteration regime activated when $\epsilon_{\mathrm{kmeans}}\le 0.8$. Noisy-CDF binning further splits each feature budget as $\epsilon_{\mathrm{hist}}=0.5\epsilon_j$, $\epsilon_{\mathrm{sum}}=0.25\epsilon_j$, and $\epsilon_{\mathrm{count}}=0.25\epsilon_j$.

\subsection{Evaluation Protocol and Metrics}

Utility is measured with the train-synthetic, test-real (TSTR) protocol using a support vector machine (SVM). For each dataset, we use 5-fold cross-validation. Thus, every parameter configuration is run five times, once per fold, and the five scores are averaged. In each fold, the real training split is discretized, synthesized by PrivMRF, and reconstructed into continuous synthetic data. The SVM is trained only on this reconstructed synthetic training set and is evaluated on the held-out real fold. This setup asks whether the synthetic data preserves decision boundaries that generalize to real data.

We report classification accuracy and macro-averaged F1 score. Accuracy measures overall predictive performance, while macro-F1 gives equal weight to each class and is therefore important for multiclass or imbalanced datasets. The result plots use three views: peak utility selects the best configuration in the declared grid, robustness summarizes fixed configurations across the grid, and severe-privacy results restrict the budget axis to $\epsilon<1.0$.

We do not use total variation distance (TVD) as a primary metric. In this pipeline, the data pass through three states: raw continuous data (A), binned data (B), and synthetic binned data from PrivMRF (C). TVD can compare only (B) and (C), because both have the same discrete support. It does not measure how much information was lost when converting (A) to (B). This matters because very coarse bins can make TVD look good while destroying useful continuous structure. TSTR is therefore a more direct end-to-end metric for this study.

%\subsection{Implementation Details}

%The DP K-Means variant follows the perturbed Lloyd-style clustering procedure~\cite{su2016} with the proposed low-budget iteration rule. The noisy-CDF variant uses Laplace-noised micro-bin counts, convolutional smoothing, isotonic regression, adaptive equal-frequency boundaries, and private bin centers for reconstruction. After either discretizer is applied, the discrete data are passed to the same PrivMRF synthesizer. All reported utility values are averaged over the same 5-fold splits so that differences reflect the discretization method rather than changes in the train-test partition.

\section{Experimental Results}

This section reports the TSTR utility of the synthetic data produced by each discretization strategy. The results are organized by metric: Fig.~\ref{fig:accuracy_results} summarizes classification accuracy, and Fig.~\ref{fig:f1score_results} summarizes macro-averaged F1 score. In both figures, the rows show three complementary views of performance: the best achievable utility within the tested parameter grid, robustness under fixed parameter choices, and the strict-privacy regime with $\epsilon<1.0$.

\subsection{Classification Accuracy}

\begin{figure}[t]
    \centering
    \begin{minipage}{0.32\textwidth}
        \centering
        \includegraphics[width=\textwidth]{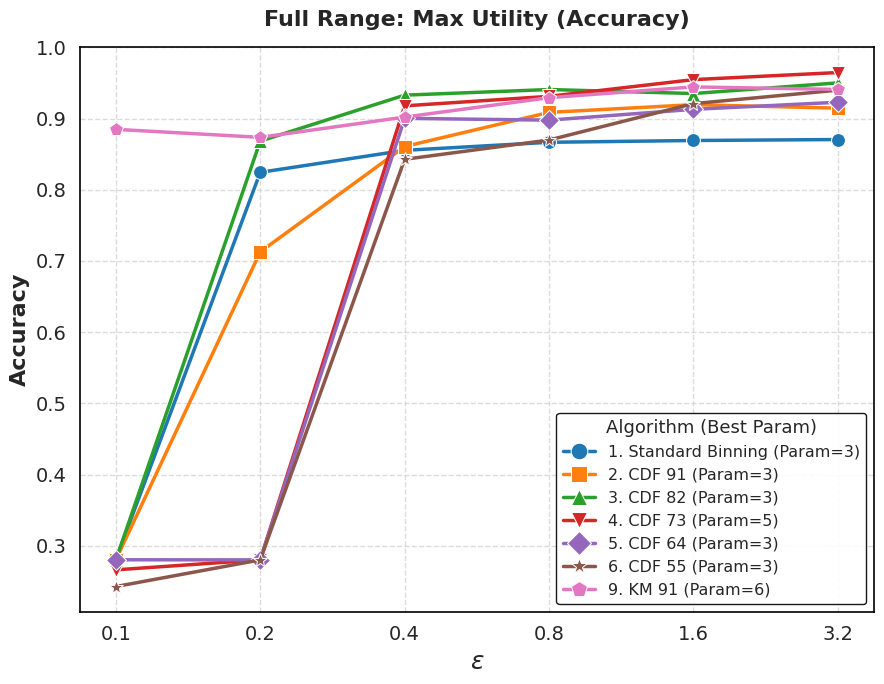}\\
        (a) Peak, Dry Bean
    \end{minipage}
    \hfill
    \begin{minipage}{0.32\textwidth}
        \centering
        \includegraphics[width=\textwidth]{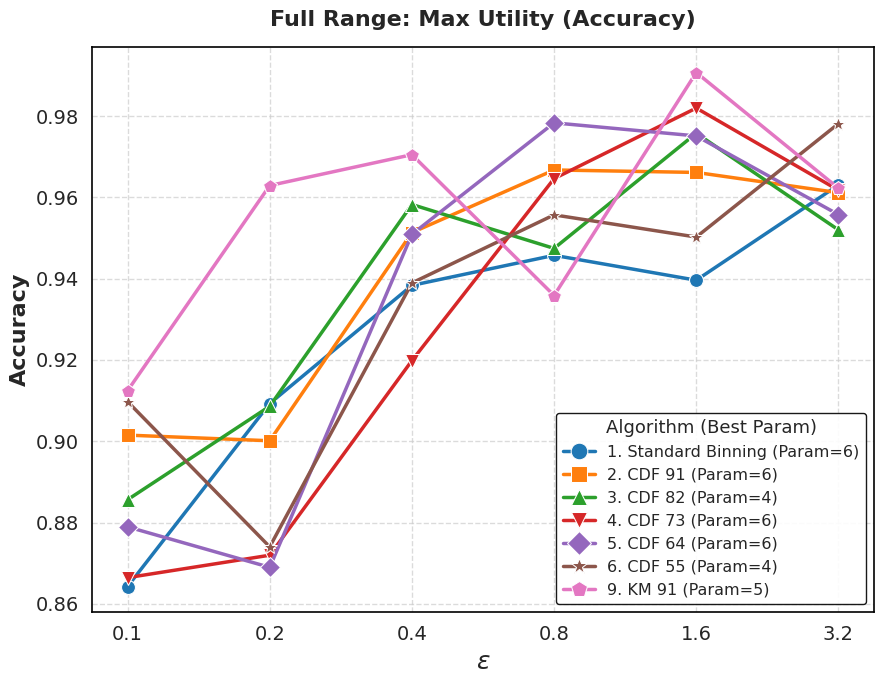}\\
        (b) Peak, MAGIC
    \end{minipage}
    \hfill
    \begin{minipage}{0.32\textwidth}
        \centering
        \includegraphics[width=\textwidth]{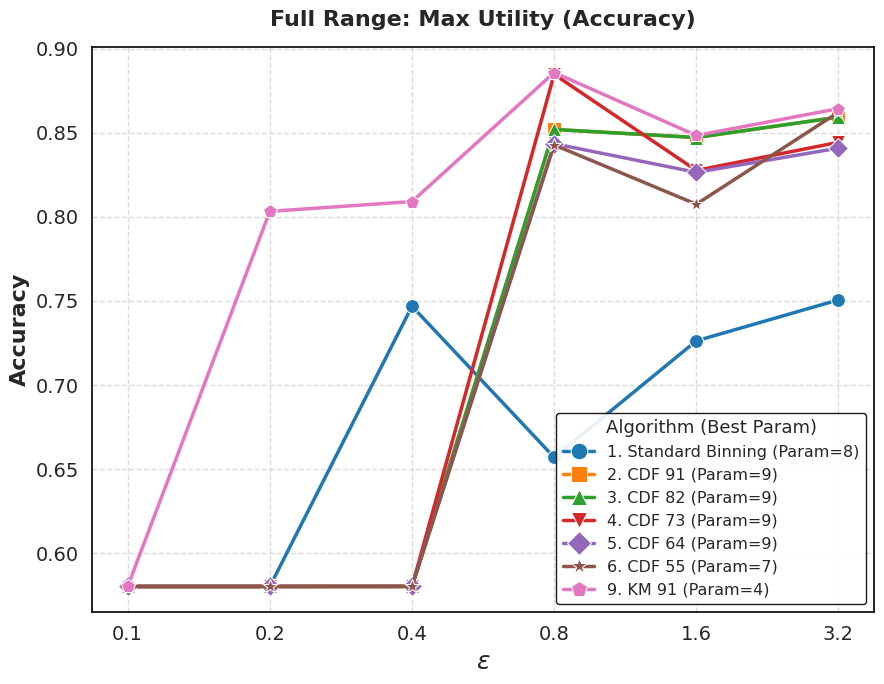}\\
        (c) Peak, Avila
    \end{minipage}
    \par\vspace{0.4em}
    \begin{minipage}{0.32\textwidth}
        \centering
        \includegraphics[width=\textwidth]{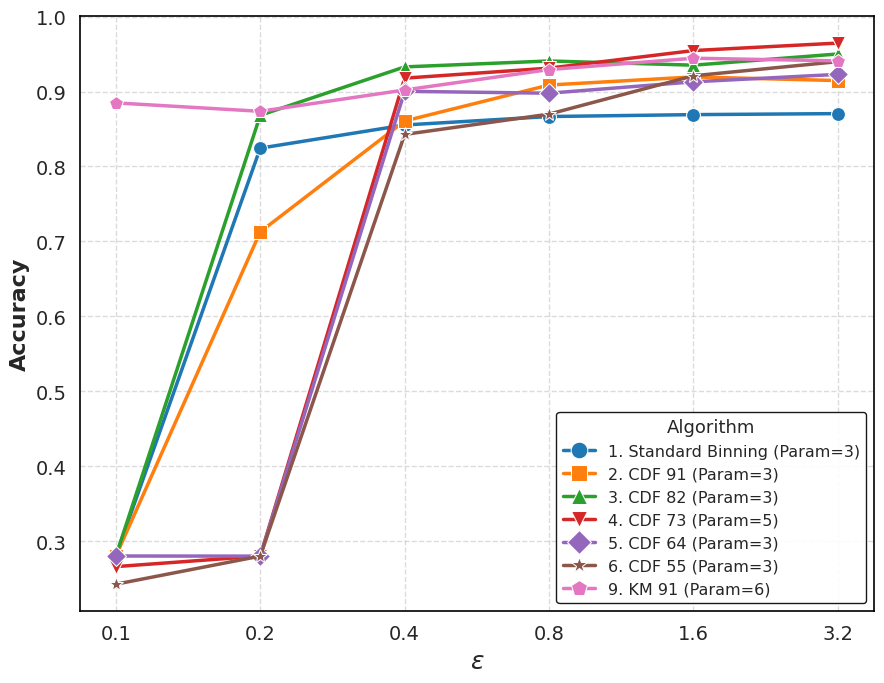}\\
        (d) Robustness, Dry Bean
    \end{minipage}
    \hfill
    \begin{minipage}{0.32\textwidth}
        \centering
        \includegraphics[width=\textwidth]{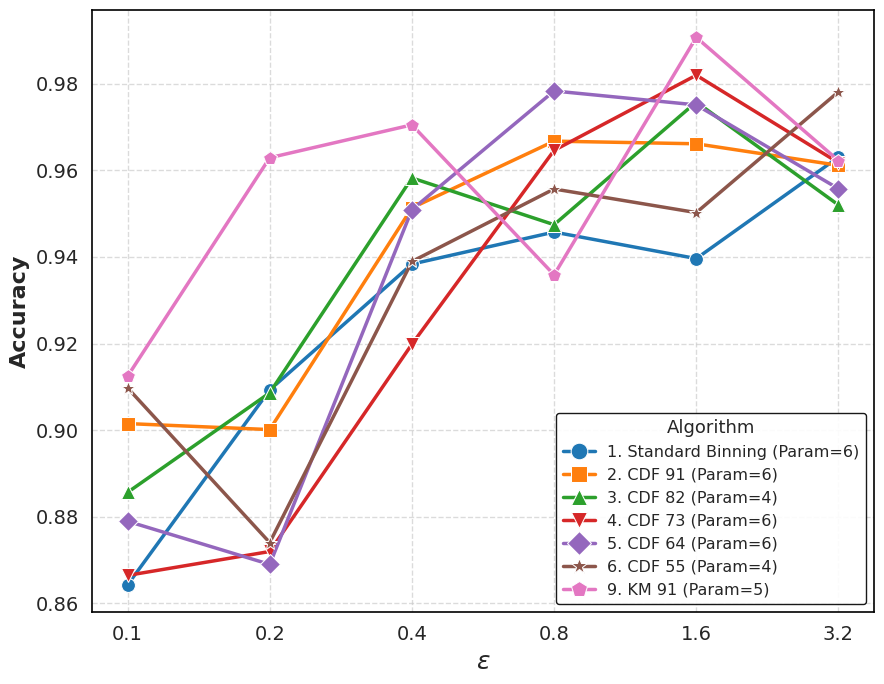}\\
        (e) Robustness, MAGIC
    \end{minipage}
    \hfill
    \begin{minipage}{0.32\textwidth}
        \centering
        \includegraphics[width=\textwidth]{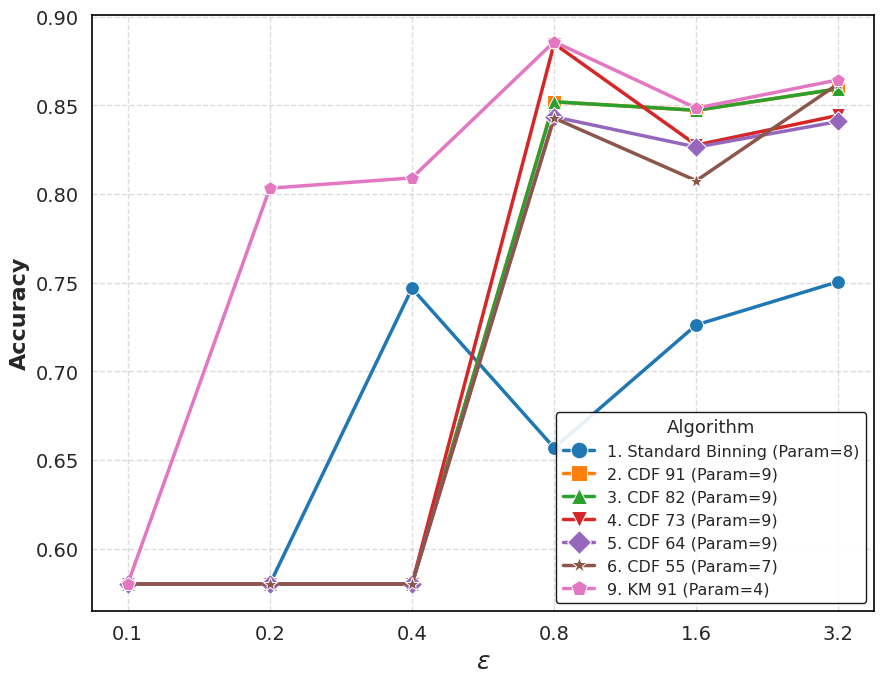}\\
        (f) Robustness, Avila
    \end{minipage}
    \par\vspace{0.4em}
    \begin{minipage}{0.32\textwidth}
        \centering
        \includegraphics[width=\textwidth]{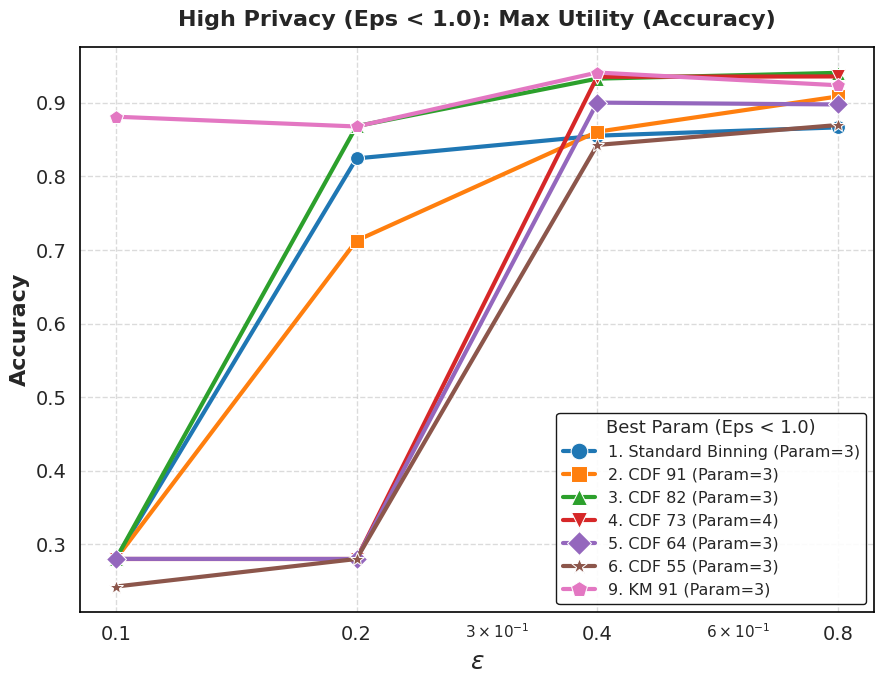}\\
        (g) Severe privacy, Dry Bean
    \end{minipage}
    \hfill
    \begin{minipage}{0.32\textwidth}
        \centering
        \includegraphics[width=\textwidth]{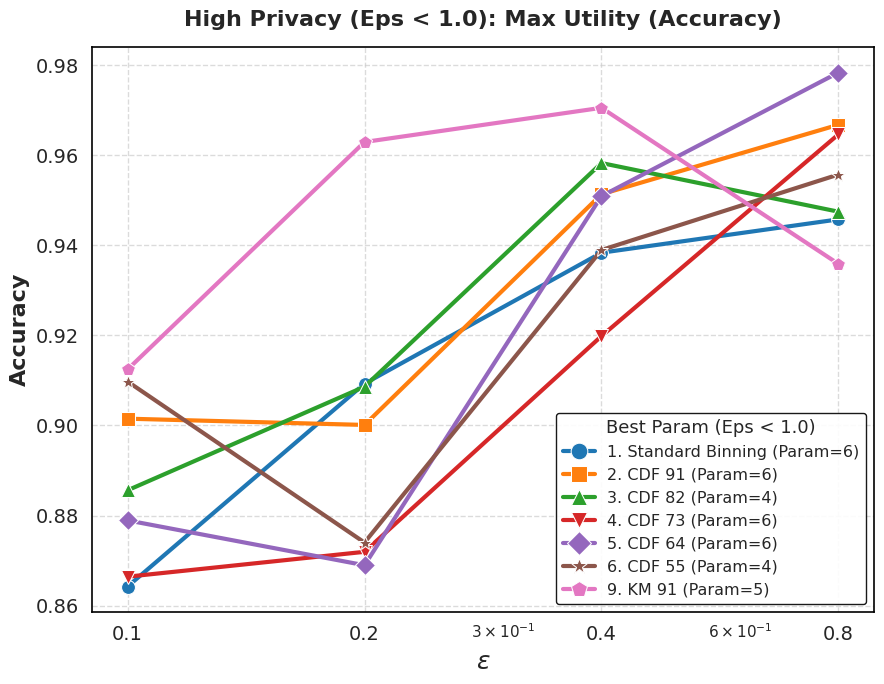}\\
        (h) Severe privacy, MAGIC
    \end{minipage}
    \hfill
    \begin{minipage}{0.32\textwidth}
        \centering
        \includegraphics[width=\textwidth]{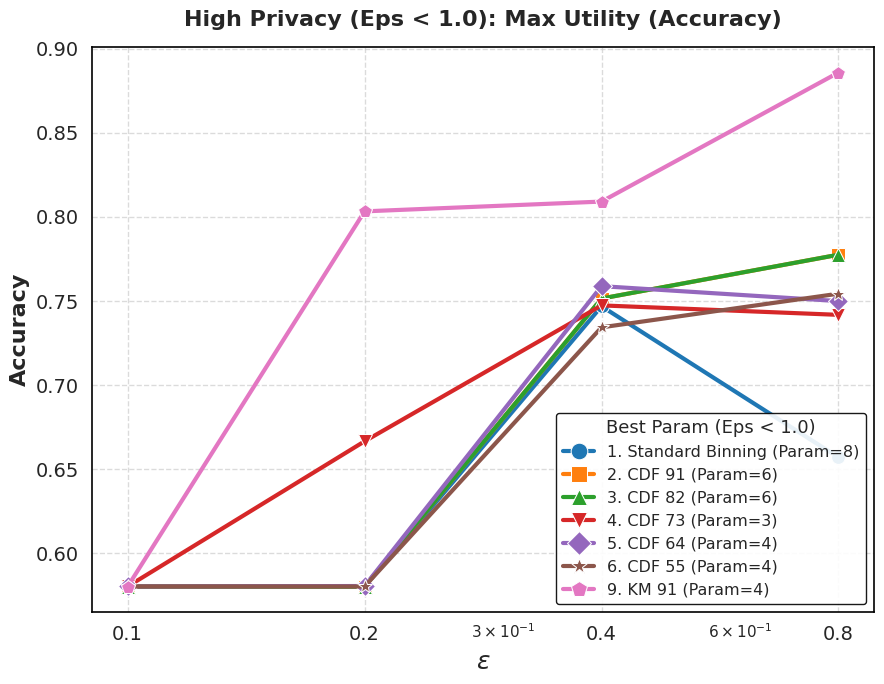}\\
        (i) Severe privacy, Avila
    \end{minipage}
    \caption{Classification accuracy results. Rows show peak achievable utility, robustness under fixed parameter configurations, and severe privacy constraints ($\epsilon < 1.0$); columns show Dry Bean, MAGIC Gamma Telescope, and Avila.}
    \label{fig:accuracy_results}
\end{figure}

\paragraph{Empirical Observations (Accuracy):}
Fig.~\ref{fig:accuracy_results} shows that accuracy depends on the combined effect of privacy noise, bin resolution, and dataset structure. As the privacy budget increases, the peak-utility curves generally improve because both the discretization step and PrivMRF receive less noisy statistics. The proposed distribution-aware discretizers are most beneficial when the privacy budget is small, where equal-width bins create sparse categories that are easily dominated by DP noise.

On Avila at $\epsilon=0.2$, the best adaptive CDF setting reaches an accuracy of 0.7158, compared with 0.6345 for \textbf{Uniform-PrivMRF}. On MAGIC at the same privacy budget, CDF~64 reaches 0.9594, while the uniform baseline reaches 0.9144. The strongest low-budget gain appears on Dry Bean: at $\epsilon=0.1$, KM~55 maintains an accuracy of 0.8848, whereas the uniform and CDF configurations drop to around 0.2806. These results indicate that private, distribution-aware binning can preserve useful decision boundaries even when PrivMRF must learn from heavily perturbed marginal statistics.

The robustness plots also show that more bins are not always better. Larger domains can represent the continuous data more finely, but they also spread private counts across more cells and increase sparsity in the graphical model. DP K-Means performs well when its private centroids remain stable, reaching 0.9420 on Avila and 0.9916 on MAGIC at $\epsilon=3.2$. The noisy-CDF method is more stable in several strict-privacy cases because smoothing and isotonic repair reduce irregularities in the noisy cumulative distribution before the bin boundaries are selected. Across all datasets, the univariate discretization strategy keeps the effective attribute domains compact enough for PrivMRF inference to complete without memory exhaustion or domain-size explosion.

\subsection{Macro-averaged F1 Score}

\begin{figure}[t]
    \centering
    \begin{minipage}{0.32\textwidth}
        \centering
        \includegraphics[width=\textwidth]{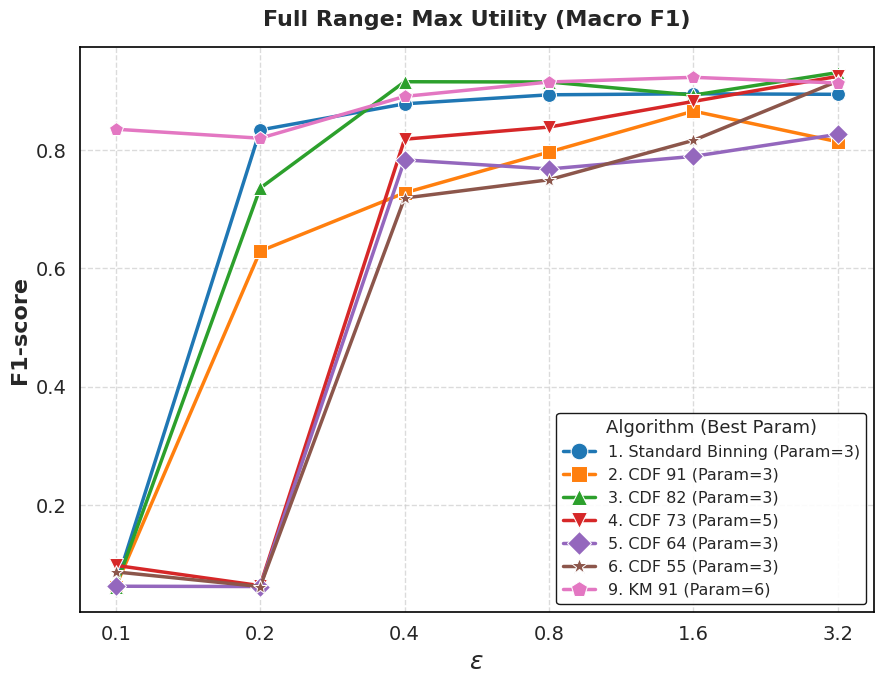}\\
        (a) Peak, Dry Bean
    \end{minipage}
    \hfill
    \begin{minipage}{0.32\textwidth}
        \centering
        \includegraphics[width=\textwidth]{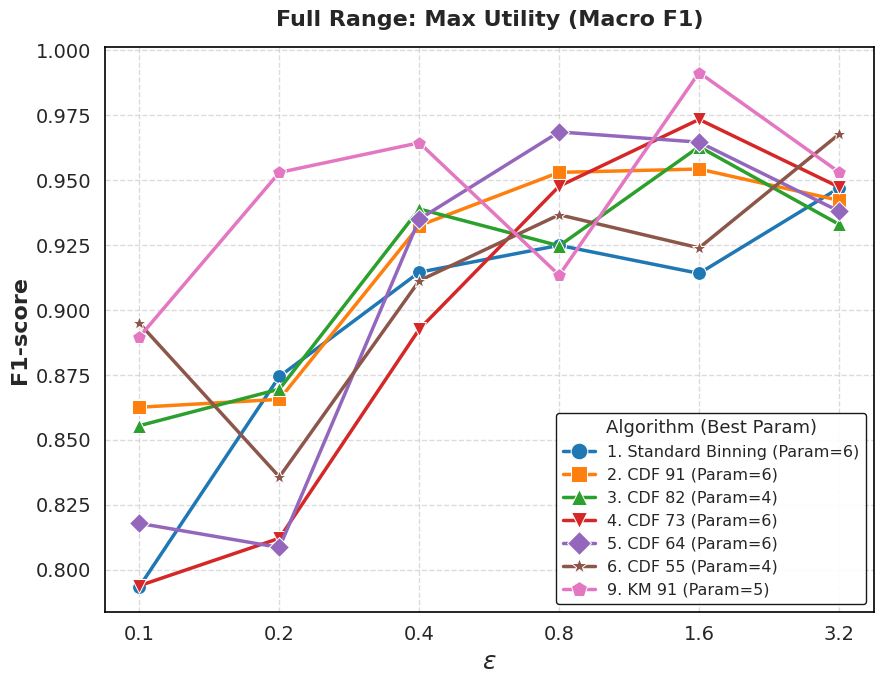}\\
        (b) Peak, MAGIC
    \end{minipage}
    \hfill
    \begin{minipage}{0.32\textwidth}
        \centering
        \includegraphics[width=\textwidth]{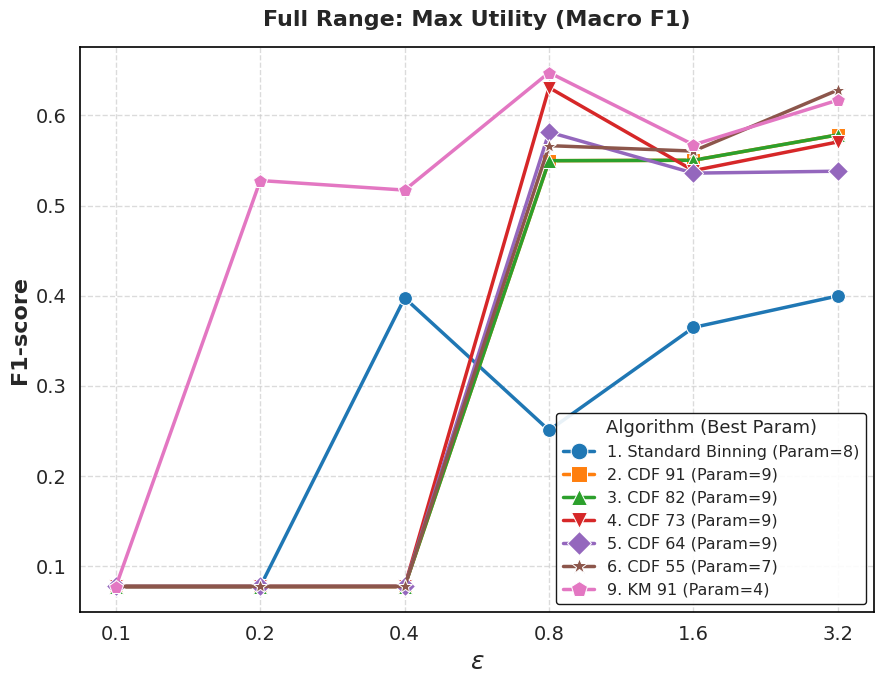}\\
        (c) Peak, Avila
    \end{minipage}
    \par\vspace{0.4em}
    \begin{minipage}{0.32\textwidth}
        \centering
        \includegraphics[width=\textwidth]{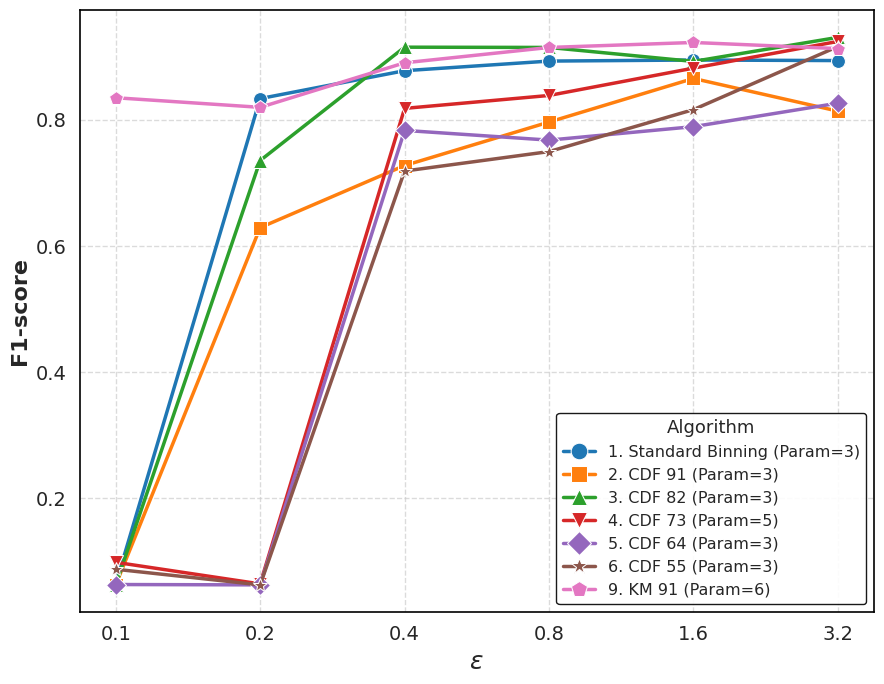}\\
        (d) Robustness, Dry Bean
    \end{minipage}
    \hfill
    \begin{minipage}{0.32\textwidth}
        \centering
        \includegraphics[width=\textwidth]{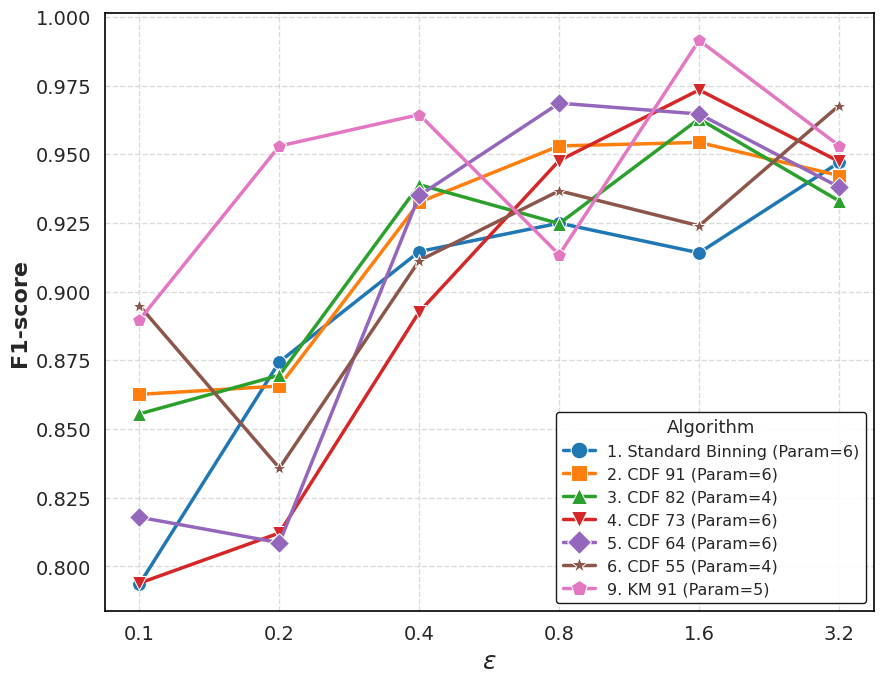}\\
        (e) Robustness, MAGIC
    \end{minipage}
    \hfill
    \begin{minipage}{0.32\textwidth}
        \centering
        \includegraphics[width=\textwidth]{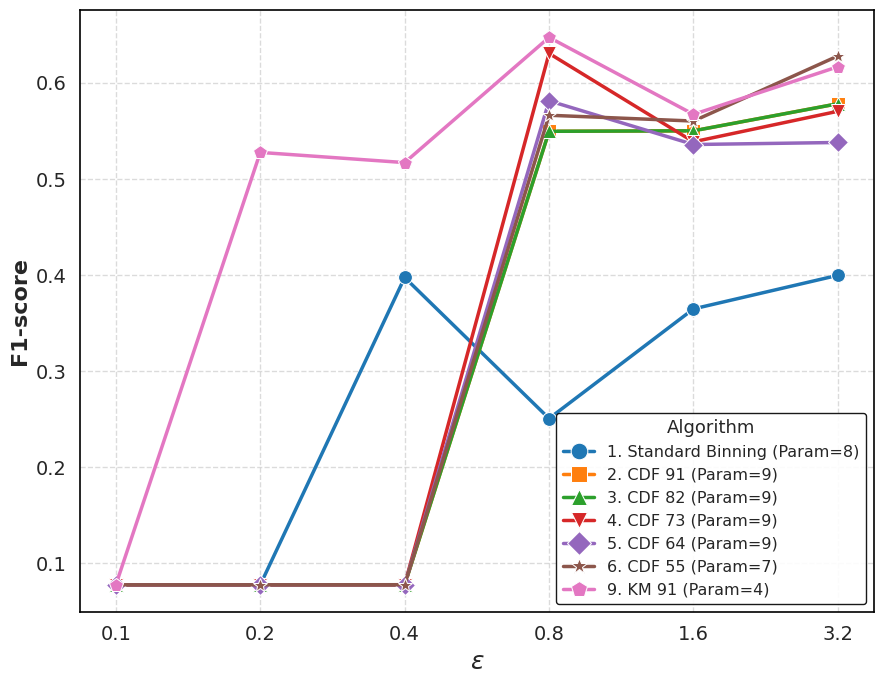}\\
        (f) Robustness, Avila
    \end{minipage}
    \par\vspace{0.4em}
    \begin{minipage}{0.32\textwidth}
        \centering
        \includegraphics[width=\textwidth]{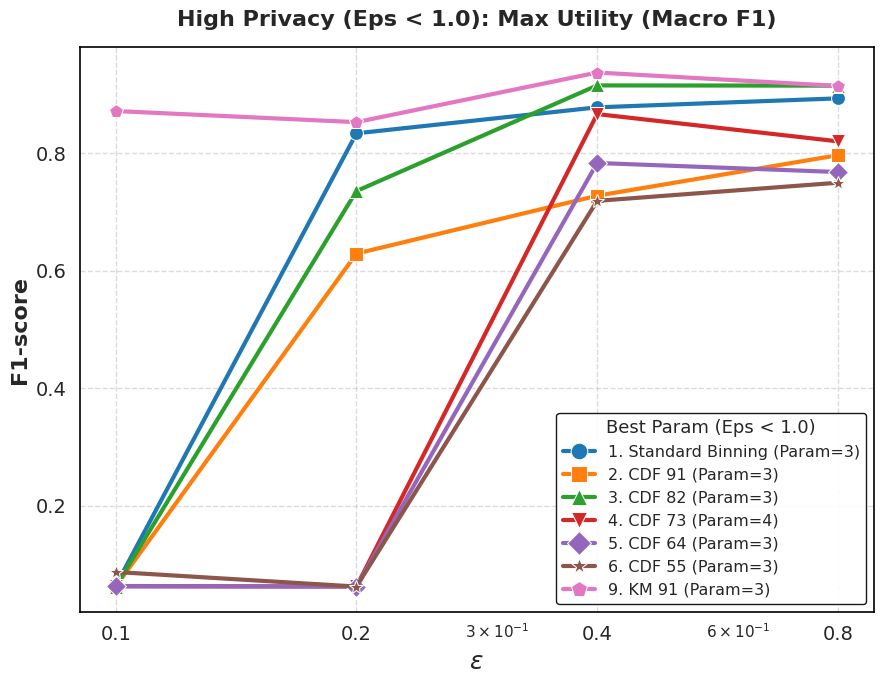}\\
        (g) Severe privacy, Dry Bean
    \end{minipage}
    \hfill
    \begin{minipage}{0.32\textwidth}
        \centering
        \includegraphics[width=\textwidth]{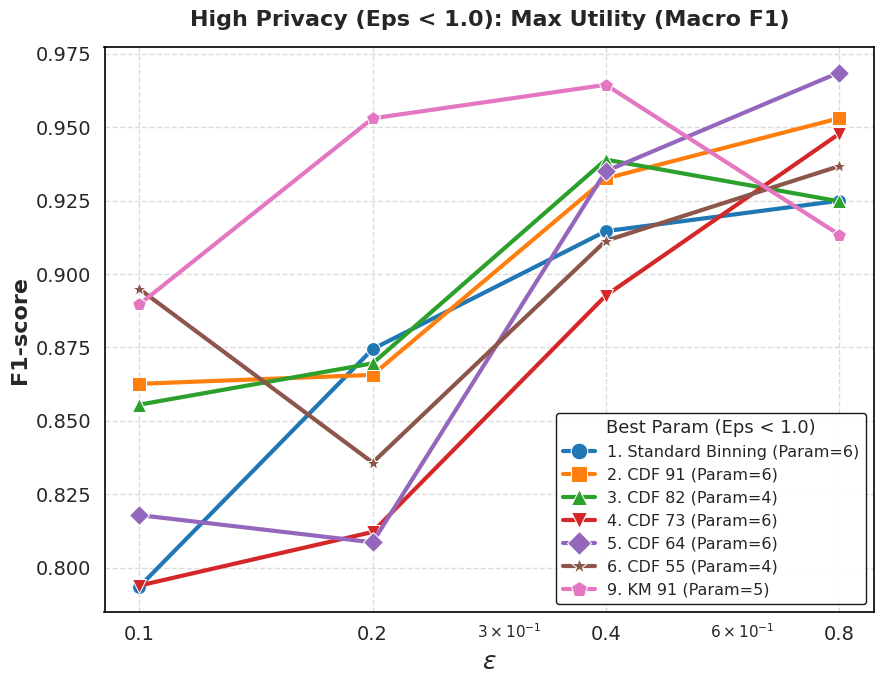}\\
        (h) Severe privacy, MAGIC
    \end{minipage}
    \hfill
    \begin{minipage}{0.32\textwidth}
        \centering
        \includegraphics[width=\textwidth]{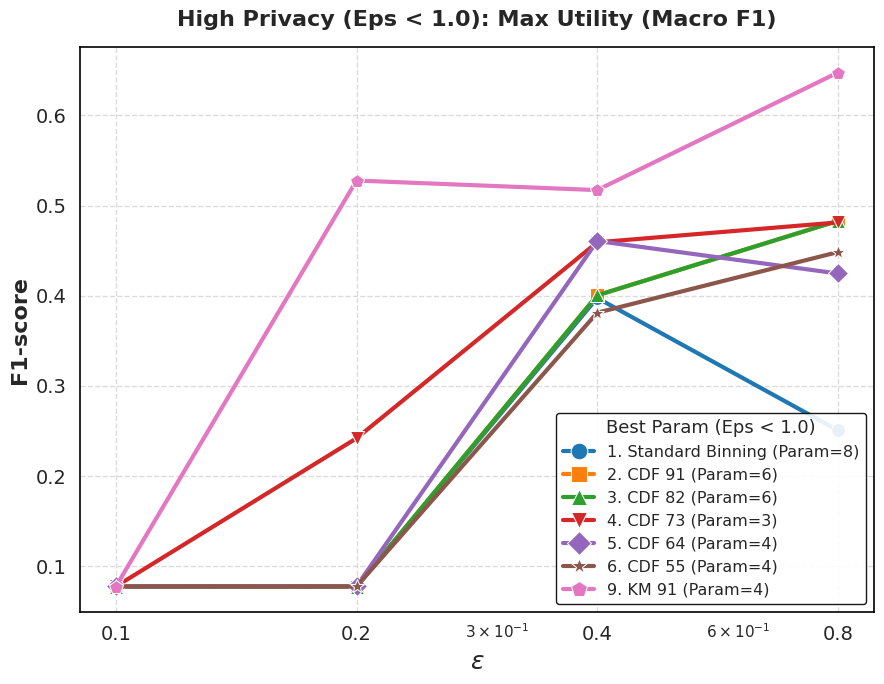}\\
        (i) Severe privacy, Avila
    \end{minipage}
    \caption{Macro-averaged F1-score results. Rows show peak achievable utility, robustness under fixed parameter configurations, and severe privacy constraints ($\epsilon < 1.0$); columns show Dry Bean, MAGIC Gamma Telescope, and Avila.}
    \label{fig:f1score_results}
\end{figure}

\paragraph{Empirical Observations (Macro-F1):}
Macro-F1 gives equal weight to each class and is therefore more sensitive to class-level failures than accuracy. This is important for Dry Bean and Avila, where minority or overlapping classes can be hidden by aggregate accuracy. Fig.~\ref{fig:f1score_results} shows that adaptive discretization improves balanced predictive utility by preserving more informative one-dimensional structure before PrivMRF models cross-feature dependencies.

On Avila at $\epsilon=0.8$, CDF~73 achieves a macro-F1 score of 0.6308, compared with 0.4298 for the equal-width binning. Under stricter privacy, the contrast is even sharper. On Dry Bean at $\epsilon=0.1$, the uniform baseline falls to a macro-F1 score of 0.0651, while KM~55 reaches 0.8718. On Avila at $\epsilon=0.2$, KM~55 obtains 0.5278, compared with 0.1709 for uniform binning. These results show that adaptive discretization can protect class boundaries that would otherwise be lost when noisy counts are spread across poorly matched equal-width bins.

The macro-F1 results also clarify the different strengths of the two proposed mechanisms. DP K-Means can produce strong class separation when the learned centroids are reliable, especially in very low-budget cases where a small number of representative values is preferable to many noisy intervals. Noisy-CDF binning provides steadier performance across moderate budgets because its smoothed and monotone CDF estimate reduces empty, reversed, or unstable intervals. Overall, the F1-score analysis confirms that useful private synthetic data requires not only high average accuracy but also discretization that preserves class-specific structure before synthesis.

\subsection{Discussion}

The results show that discretization is not merely a preprocessing detail when PrivMRF is applied to continuous data. Because PrivMRF estimates noisy low-dimensional marginals over finite domains, the choice of bins directly controls both information preservation and the sparsity seen by the graphical model. Equal-width binning is simple, but it ignores density variation; in skewed or imbalanced data, it can allocate many categories to low-density regions and too few categories to dense regions. The proposed methods improve utility by making the categorical representation depend on the private marginal distribution of each feature.

The two proposed mechanisms are useful in different regimes. DP K-Means binning uses private centroids as representative values and can give strong performance when a small set of stable centers captures the feature distribution. This behavior is especially helpful under severe privacy budgets, where many interval boundaries would produce noisy and sparse categories. However, iterative centroid updates repeatedly use noisy statistics, so DP K-Means can degrade when the per-feature budget is too small. The budget-adaptive iteration limit mitigates this risk by prioritizing stability over repeated convergence steps.

Noisy-CDF adaptive binning, by contrast, focuses on stable boundary estimation. It uses a private histogram to estimate the marginal cumulative distribution and then repairs the noisy curve before selecting quantile-like boundaries. Smoothing reduces local variance, while isotonic regression enforces the monotonicity required of a valid CDF. This makes the method less sensitive to isolated noisy micro-bins and explains its strong accuracy and macro-F1 performance across several low- and moderate-budget settings.

A practical implication is that the best configuration is not determined only by maximizing the number of bins. Finer bins can preserve more continuous detail, but they also enlarge the domains on which PrivMRF must estimate noisy marginals. Under strict privacy constraints, compact and stable bins can be more useful than highly expressive but sparse ones. The framework therefore separates the problem into two manageable parts: private univariate discretization controls the continuous-to-categorical representation, and PrivMRF then models dependencies among the resulting categorical attributes without requiring joint discretization over the original continuous space.

\section{Conclusion and Future Works}

This paper studied differentially private synthetic data generation for continuous tabular data using PrivMRF. Since PrivMRF requires categorical input, the utility of the final synthetic data depends strongly on how continuous attributes are discretized before synthesis. We proposed two PrivMRF-compatible univariate discretization strategies: DP K-Means binning with a budget-adaptive iteration rule, and noisy-CDF adaptive binning with smoothing, isotonic repair, and private representative centers. After PrivMRF generates synthetic discrete records, these records are reconstructed into continuous form and evaluated using downstream SVM classification; this evaluation is post-processing of the released synthetic data and therefore does not consume additional privacy budget.

The experiments on Dry Bean, MAGIC Gamma Telescope, and Avila show that distribution-aware discretization can substantially improve TSTR utility over equal-width binning, especially under strict privacy constraints. Adaptive CDF binning improves accuracy on Avila and MAGIC at $\epsilon=0.2$, while DP K-Means gives a large macro-F1 improvement on Dry Bean at $\epsilon=0.1$. These results support the central conclusion that private synthetic data generation for continuous domains benefits from treating discretization as part of the private mechanism rather than as a fixed preprocessing step.

Several directions remain for future work. In the current experiments, the budget split and number of bins are selected from fixed grids, and each continuous feature receives the same binning budget. Future studies could allocate privacy budgets and bin resolutions adaptively across features or combine the stability of univariate discretization with limited multivariate boundary refinement. A further limitation is that this study isolates the effect of discretization within a fixed PrivMRF pipeline and therefore does not compare the proposed interfaces with end-to-end continuous DP generators such as PATE-GAN and DP-CTGAN. A broader evaluation across these different model families is also an important direction for future work.

%\paragraph{Code availability.}
%The implementation is available at \url{https://github.com/koki-223/PrivMRF-Adapted.git}.

\section*{Acknowledgments}
We would like to thank the Faculty of Science, Chiang Mai University, Thailand, for its financial support. This research was also facilitated by the University of Tokyo, Japan. Computational resources were provided by the Information Technology Service Center (ITSC) and the Chiang Mai University High Performance Computing (HPC) facility. Vorapong Suppakitpaisarn's research work has been supported by KAKENHI Grant JP25K00369.

\bibliographystyle{splncs04}
\bibliography{references}

\end{document}